\documentclass{article}
\usepackage{acra}

\usepackage{times}
\usepackage{amssymb}

\usepackage{graphics} 
\usepackage{epsfig} 
\usepackage{amsmath} 
\usepackage{multirow}
\usepackage{tabularx}
\usepackage{booktabs}
\usepackage{tabulary}
\usepackage{graphicx}
\usepackage{footnote}

\usepackage{numprint}
\usepackage{collcell}
\usepackage[table]{xcolor}
\usepackage{xstring}
\usepackage{pgfplots}

\usepackage[T1]{fontenc}
\usepackage[english]{babel}

\usepackage{caption}

\author{
	Andrew Spek \\ Monash University, Australia \\ andrew.spek@monash.edu \\ 
	\And 
	Wai Ho li \\ Monash University, Australia \\ wai.ho.li@monash.edu \\
	\And 
	Tom Drummond \\ Monash University, Australia \\ tom.drummond@monash.edu \\
}
\title{A Fast Method For Computing Principal Curvatures From Range Images}

\begin{document}

\maketitle

\begin{abstract}
	
	Estimation of surface curvature from range data is important for a range of tasks in computer vision and robotics, object segmentation, object recognition and robotic grasping estimation. This work presents a fast method of robustly computing accurate metric principal curvature values from noisy point clouds which was implemented on GPU. In contrast to existing readily available solutions which first differentiate the surface to estimate surface normals and then differentiate these to obtain curvature, amplifying noise, our method iteratively fits parabolic quadric surface patches to the data. Additionally previous methods with a similar formulation use less robust techniques less applicable to a high noise sensor. We demonstrate that our method is fast and provides better curvature estimates than existing techniques. In particular we compare our method to several alternatives to demonstrate the improvement. 
	
\end{abstract}


\section{Introduction}

\begin{figure}[ht!]
	\begin{center}
		\includegraphics[width=.8\linewidth]{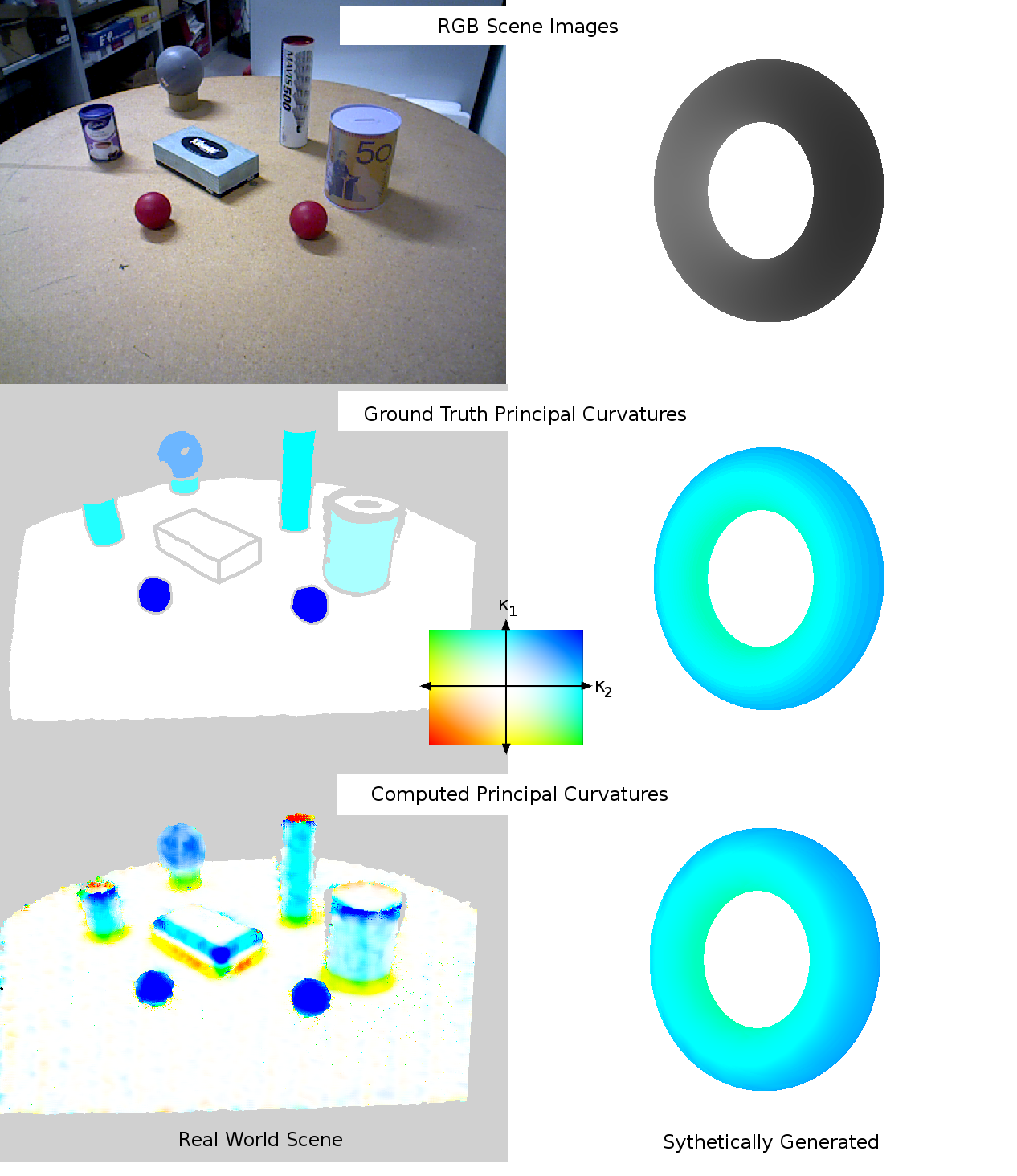}
		\caption{Example images from the evaluation datasets created for testing real-world and synthetic scenes. The rows show in order, an rgb image of the scene, the ground truth principal curvature values and the computed principal curvature values using our method. The key used for colouring the principal curvatures is shown center.}
		\label{fig:dataset_light}
	\end{center}
\end{figure}

Curvature is a robust geometric surface feature that is viewpoint invariant which makes it useful for various computer vision and robotics applications. Surface curvature in this case is the principal curvatures as opposed to Gaussian and mean curvatures of an object. The principal curvature is the rate at which the surface normal angle changes as you move along the surface, both maximally and minimally. This is a useful quantity that can be computed from meshes, point clouds and more recently RGB-D data. Surface curvature has been used in computer graphics for applications such as model de noising and surface registration \cite{Mitra2004}. Surface curvature has also been used in robotics for scene/object segmentation \cite{Besl1988} \cite{Guillaume2004a} \cite{Lee2014}. One important quality of curvature is its viewpoint invariant nature, that is the curvature of a surface appears the same regardless of the view you observe the surface from. 

Many of the recent methods for computing surface curvature generally assume the data used will be of high quality \cite{Abdelmalek1990} \cite{Griffin2011} \cite{Griffin2012} \cite{Rusinkiewicza}, and compute principal curvature using double differentiation of the surface or approximations on very small sections of the surface. These methods of computing surface curvature are highly susceptible to noise. This has led many techniques to either smooth the surface representation \cite{Marton2011} in order to improve the resulting curvature estimates or using more robust statistical techniques such as re-weighted least-squares \cite{Besl1988}. The introduction of low cost RGB-D sensors such as the Microsoft Kinect and ASUS XTion, has provided a fast method of collecting relatively high resolution dense range images at frame rate for a consumer market. These sensors use known pseudo-random projected dot pattern and measure the scene geometry to interpolate a depth value for each pixel in a VGA sized image. Khoshelham et. al. demonstrated in \cite{Khoshelham2012} the data produced is very noisy, making it difficult to measure curvature accurately using these low-noise methodologies. The current iteration of the Kinect (for Xbox One) uses a time-of-flight camera which measures depth by integrating the returned infra-red light. This sensor was also tested using our method and showed slightly lower levels of noise compared to the original Kinect, but the overall depth estimates were less reliable.


Many applications in computer vision and robotics have been expensive to compute in real-time and techniques have been developed to overcome the computational limitations by processing data selectively. A growing trend is to instead use General Purpose GPU (GPGPU) programming which utilising the large throughput of the GPUs that have been built for graphics applications to perform large numbers of computations in parallel. Lee et. al. demonstrated in \cite{Lee2014} that this enables real-time segmentation and curvature estimation of large scale point clouds. However their techniques still require preprocessing steps which are not real-time.

Point Cloud Library is a widely used open-source library for processing most forms of model data including point clouds and meshes, presented by Rusu et. al. in \cite{Rusu2011}. PCL provides multiple methods of computing an estimated curvature value directly from point cloud data, with CPU and GPU implementations. However the estimated metric curvature value is estimated based on a PCA of the change in surface normals (which are also estimated using a PCA of surface changes), and although this is a smoothing method, it can still be greatly effected by surface noise especially when taken to second order. PCL also provides the option to pre-process the data to smooth the data but this again leads to additional computations and the creation of new data. Additionally many of the quantities labelled surface curvature are actually the surface variation described by Pauly et al. in \cite{Pauly2002} which is a quantity that is proportional to the mean absolute curvature but non-metric. 

The key contribution of this paper is a real-time GPU based algorithm for estimating metric curvature values from range images via an iterative quadric surface patch fitting. This method provides several benefits:
\begin{itemize}
	\item It provides more accurate metric estimates of real-world curvature values than existing methods demonstrated in section \ref{sec:system_accuracy}.
	\item It provides smoother and more accurate surface normal estimates compared to surface differentiation by PCA shown in section \ref{sec:normal_est}
	\item The method is fast and easily able to run at frame-rate as shown in section \ref{sec:timing}
	\item The metric curvature estimates produced by our system can be used to accurately estimate object correspondences across multiple viewpoints as shown in section \ref{sec:curv_correspond}
	\item It works well with noisy point cloud data, such as that produced by low-cost RGB-D sensors (like the Microsoft Kinect and ASUS XTion).
	\item Open-source implementation\footnote{https://github.com/aspek1/QuadricCurvature}
\end{itemize}


\section{Related Work}

There exists a significant body of research concerning the computation of surface curvature \cite{Besl1988}\cite{Abdelmalek1990}\cite{Gatzke2006}\cite{Griffin2012}. Many of these methods are computed using dense point cloud data \cite{Besl1988} and high quality meshes \cite{Griffin2011}\cite{Gatzke2006}\cite{Griffin2012}\cite{Rusinkiewicza} allowing high quality estimates of curvature to be made using a relatively small amount of the data, due to the high inlier rates. In \cite{Besl1988} Besl introduces the concept of local surface fitting in order to compute curvature. Besl et al examines fitting range image data densely using least-squares regression to fit to quadratic, cubic and quartic surface parametrizations. Surface fitting has been a popular choice for surface curvature estimation \cite{Flynn1989a}\cite{Alshawabkeh2008}, in particular quadric surface fitting which has been shown to be effective for curvature based segmentation \cite{Guillaume2004a}. The system produced for this paper is most similar to the work in \cite{Douros2002}, in that both attempt to fit local surface patches to point cloud data directly, however in their work Douros et al. use a least squares approximation to produce a closed form solution. Where as our method fits a surface iteratively based on a quadric patch, that directly attempts to minimise error in principal curvature estimation. We compare the performance of this formulation to our method and several others in section \ref{sec:results}.

Curvature estimation from quadric surface fitting is based on fitting an implicit quadric surface around a central point. Curvature is then estimated from calculated surface gradient estimations of this fitted surface. In order to estimate surface curvature some methods \cite{Rusu2010a}\cite{Rusu2011} also employ a technique that requires the estimation of second order partial surface derivatives about the point. This creates an estimate highly susceptible to noise in the point cloud. 

Many common methods employ techniques that require high quality surface meshes or point clouds to estimate normals and curvature fast and accurately. In \cite{Rusinkiewicza} Rusinkiwicz et al. use a patch of connected vertices taken around each vertex in a triangular mesh to estimate curvature. The patch used is made of a 1-ring selection of triangles that share a vertex with that point. Using this small neighbourhood of triangles, normals are estimated for each vertex and  their relative weight is computed based on the proportion of the area of the surface that lies closest to the vertex. Using this 1-ring of normals all pair-wise normal differences are calculated and the curvature of the surface is estimated from the weighted sum of these normal differences. This method is roughly equivalent to a discrete differentiation of the surface normal field and gives a reportedly reliable estimate of the principal surface curvatures across the mesh. However as the patches used in this 1-ring method only uses direct neighbours to the center vertex it is heavily sensitive to noise in the surface. This method was extended in \cite{Griffin2012}, where Griffin et al. used a GPU implementation to make it run in realtime (30fps) for mesh models of the order a million vertices. However this requires a pre-processing step to sort the vertices via an iso-surface extraction which takes on the order of hundreds of ms (using NVIDIA GTX480) for models of that size. 

There also exist other open-source methods for computing curvature \cite{Rusu2011}, namely the method developed for the open-source Point Cloud Library (PCL) by Rusu et al. This library provides an estimate for surface curvature in a point-cloud and can be applied generically to any point-cloud. The original method employs a KD-tree which allows a metric based surface patch which is proportional to the density of the point cloud, making it suitable for noisy point clouds. They also created a GPU method which operates in real-time and operates in view-space (as our method does), with both computing an estimate for principal curvature values given an input point cloud. However both methods compute curvature via double differentiation using PCA of the surface differences and then normal differences. This double differentiation substantially amplifies the noise in the point cloud data and is therefore unsuitable for noisy data sources such as Kinect or Xtion.


\section{Background and Method}

\subsection{Theory}

\subsubsection{Curvature}

The curvature of a two dimensional surface embedded in three
dimensions is defined by the rate at which a unit surface normal
changes with respect to motion across the surface.  For a twice
differentiable surface $M$, let $\hat{u}$ and $\hat{v}$ be orthogonal
unit vectors in the tangent plane to the surface at a point $p$.  The
surface normal is then given by $\hat{n}=\hat{u}\wedge\hat{v}$.
$\hat{u}$, $\hat{v}$ and $\hat{n}$ form an orthonormal basis about
$p$.  The surface in the neighbourhood of $p$ can then be defined as
\begin{equation*}
m = p + x \hat{u} + y\hat{v} + z \hat{n}
\end{equation*} 
where $z=z(x,y)$.  The surface normal can then also be defined in the
neighbourhood of $p$ as $\hat{n}(x,y)$.

The curvature of the surface is defined in terms of the second
fundamental form which is a symmetric two-form matrix which can be
represented by a $2 \times 2$ matrix in the
surface coordinate frame defined by $\hat{u}$ and $\hat{v}$:
\begin{equation}
\label{equ-curv_tensor}
\mathbf{II} = 
\left(
\begin{array}{cc}
du & dv \\
\end{array}
\right)
\left[
\begin{array}{cc}
A & B \\ B & C \\
\end{array}
\right]
\begin{pmatrix} du \\ dv \end{pmatrix}
\end{equation}
where $A$, $B$ and $C$ can be defined in terms of derivatives of the
surface normal:
\begin{align*}
A & = - \frac{\partial \hat{n}(x,y)}{\partial x}\cdot \hat{u} \\
B & = - \frac{\partial \hat{n}(x,y)}{\partial x}\cdot \hat{v}
= - \frac{\partial \hat{n}(x,y)}{\partial y}\cdot \hat{u} \\
C & = - \frac{\partial \hat{n}(x,y)}{\partial y}\cdot \hat{v}
\end{align*}
This is the usual method used to compute the curvature.  However $A$,
$B$ and $C$ can be equivalently defined as:
\[
A = \frac{\partial^2 z(x,y)}{\partial x^2}, B = \frac{\partial^2 z(x,y)}{\partial x \partial y}, C = \frac{\partial^2 z(x,y)}{\partial y^2}
\]

Using this definition, it is easy to construct a parabolic surface
that has this curvature at the origin:
\begin{equation}
z = \frac{A}{2} x^2 + Bxy + \frac{C}{2} y^2
\label{eqn-parabolic}
\end{equation}

Computing the Eigenvalues of the symmetric matrix in $\mathbf{II}$ from Equation \ref{equ-curv_tensor} gives the principal curvature values \(\kappa_1\) and \(\kappa_2\). Diagonalising this matrix with a rotation is equivalent to aligning the surface to the principal directions:

\begin{equation}
\label{eqn-final_form}
\textbf{II} = \left[
\begin{array}{cc}
du' & dv'
\end{array}
\right]
\left[
\begin{array}{cc} 
\kappa_1 & 0 \\ 
0 & \kappa_2 \\ 
\end{array}
\right]
\left[
\begin{array}{c}
du' \\
dv' \\
\end{array}
\right]
\end{equation}

\subsubsection{Quadrics}
\label{subsec:quadrics}

A Quadric in three dimensions is the implicit surface generated by the
zeros of a quadratic function in $x$, $y$ and $z$.  This matrix can be
represented by a symmetric $4 \times 4$ matrix $Q$ acting on
homogeneous coordinates.  Thus the surface is defined by:
\begin{equation*}
\left(
\begin{array}{cccc}
x & y & z & 1 \\
\end{array}
\right)
Q
\begin{pmatrix} x \\ y \\ z \\ 1 \end{pmatrix}
\end{equation*}

For the parabolic surface defined in Equation \ref{eqn-parabolic}, $Q$
is given by:
\begin{equation}
Q = \left[ 
\begin{array}{cccc}
\tfrac{A}{2} & \tfrac{B}{2} & 0 & 0 \\ [4pt]
\tfrac{B}{2} & \tfrac{C}{2} & 0 & 0 \\ [4pt]
0 & 0 & 0 & -\frac{1}{2} \\ [4pt]
0 & 0 & -\frac{1}{2} & 0 \\
\end{array} 
\right]
\end{equation}

The curvature can then be estimated iteratively by fitting a parabolic surface of this form to a local patch of range data.  In order to do this, points in the neighbourhood of $p$ are first expressed relative to $p$:
\begin{equation}
p_i' = p_i - p
\end{equation}
This means all points in the patch are expressed relative to the center coord, similar to mean shifting the data. This isn't a requirement of the algorithm but it increases the numerical stability of the system. These points must then be rotated so that the surface normal is aligned with the $z$ axis. This rotation is parametrized by a rotation $R$ in the $x$-$y$ plane. Finally, the best fitting parabolic surface might not pass through $p$ and so translation along the $z$ axis is also permitted. This rotation and translation can be represented by a
Euclidean transformation matrix:
\begin{equation*}
E = \begin{bmatrix}
\ddots &  &   & 0 \\
& R & & 0 \\
& & \ddots & t_z \\
0 & 0 & 0 & 1
\end{bmatrix}
\end{equation*}

Then the $p$-relative points $p_i'$ should then satisfy
\begin{equation}
\label{eq:quad}
p_i'^T E^T Q E p_i' =0
\end{equation}

The principal curvatures $\kappa_1$ and $\kappa_2$ can then be
obtained from $A$, $B$ and $C$ by:

\begin{equation*}
\begin{aligned}[c]
T1 &= \frac{A+C}{2}\\
T2 &= \sqrt{T1^2 - AC + B^2}\\
\end{aligned}
\hspace{1em}
\begin{aligned}[c]
\kappa_1 &= T1+T2\\
\kappa_2 &= T1-T2\\
\end{aligned}
\end{equation*}

\subsection{System Design}

\begin{figure}[h!]
	\centering
	\includegraphics[width=0.3\textwidth]{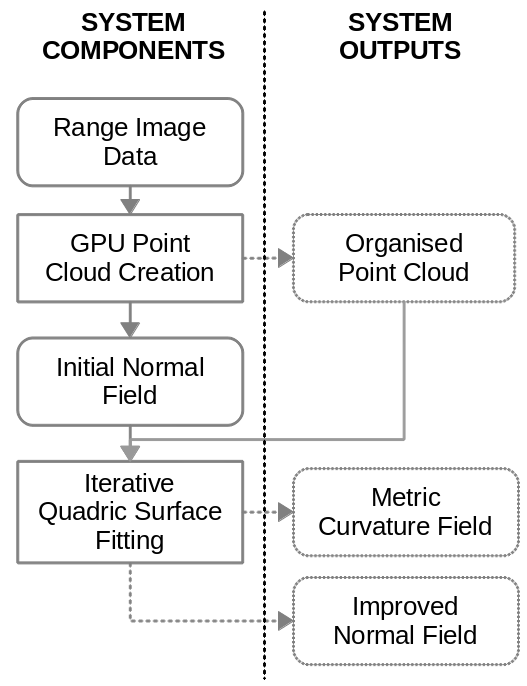}
	\caption{The operational system flow diagram for the high-level function of the metric curvature estimation system. Outputs include a point-cloud organised in raster scan order, improved normal estimates and a metric curvature field and are indicated on the left.}
	\label{fig:system-design}
\end{figure}

The following details the steps performed by our system shown in Figure \ref{fig:system-design} in computing curvature estimates from quadric surface fitting.

\begin{itemize}
	\item Point clouds are created and input; generally using RGB-D data produced by low-cost RGB-D sensors.
	\item A surface normal is computed for each point using our CUDA implementation by differentiating a \(7\times7\) patch centred on the each point.
	\item The quadric surface defined in section \ref{subsec:quadrics} is iteratively fit to an oriented \(37\times37\) patch around each point.
	\item The final quadric is used to estimate the principal curvatures
	\item The initially computed normals are refined using the computed orientation of the quadric surface.
\end{itemize}

\subsubsection{Initial Normal Estimation}

In order to compute the curvature using our method an initial estimate
of the surface normals is required. We compute this using our CUDA
implementation of a local surface differentiation, using a relatively
small \(7\times7\) patch. We simplify the calculation by parametrizing
the surface approximating $z$ as a linear function of $x$ and $y$
using regression. Where $x,y$ and $z$ are in normalized camera coordinates but could also be in normalized image coordinates. The mean values $\bar{x}$, $\bar{y}$ and $\bar{z}$ are computed
for the patch and then the matrix
\begin{equation*}
M=\left[\begin{array}{cc}
\sum_i (x_i-\bar{x})^2 & \sum_i (x_i-\bar{x})(y_i-\bar{y}) \\
\sum_i (x_i-\bar{x})(y_i-\bar{y}) & \sum_i (y_i-\bar{y})^2
\end{array} 
\right]
\end{equation*}
and the vector
\begin{equation*}
v = \left(\begin{array}{c}
\sum_i (x_i-\bar{x})(z_i-\bar{z}) \\
\sum_i (y_i-\bar{y})(z_i-\bar{z})
\end{array}
\right)
\end{equation*}
Setting \( \left( \begin{array}{cc} a & b\\ \end{array} \right) ^T = M^{-1}v \) then gives the surface normal as
\begin{equation}
\label{eq:initialnormal}
\hat{n} = \frac{1}{\sqrt{1+a^2+b^2}}
\left(
\begin{array}{c}
-a \\ -b \\ 1
\end{array}\right)
\end{equation}
and this can be cheaply computed from a
\(2\times2\) inverse matrix equation.

\subsubsection{Principal Curvature Estimatation from Quadrics}

\label{subsec:curv_quad}

\begin{figure}[h!]
	\centering
	\includegraphics[width=0.7\linewidth]{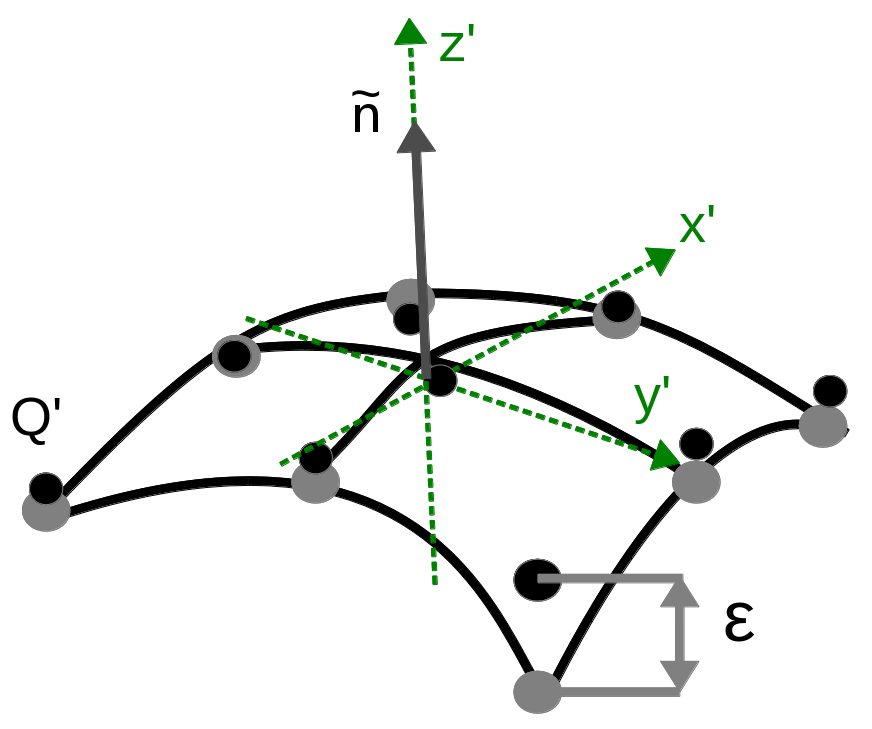}
	\caption{A geometric interpretation of fitting the quadric surface we have defined for computing principal curvature values.}
	\label{fig:quad_opt}
\end{figure}

The sum of the error function is written as:
\begin{equation}
\label{eq:error_func}
\epsilon = \sum\limits_{i=1}^n p_i'^T Q' p_i'
\end{equation}
where $Q' = E^T Q E$ and is parameterised by six parameters. These comprise three curvature parameters ($A$, $B$ and $C$), two rotation parameters ($\theta_x$ and $\theta_y$) to rotate the surface normal into the $z$ axis) and one translation parameter ($t_z$) which allows the surface to move away from the center point. Figure \ref{fig:quad_opt} shows a geometric representation of the frame around $p$ and demonstrates what is meant by the error $\epsilon_i$.

The initial surface normal estimate from equation \ref{eq:initialnormal} is used to initialise $R(\theta_x, \theta_y)$, making the $z$ axis align with the normal. The surface is initialised to be planar ($A=B=C=0$) and with zero z offset ($t_z=0$).

\subsubsection{Iterative Re-weighted Surface Fitting}
\label{subsec:iterative_reweighting}

Given the error function defined in section \ref{subsec:curv_quad} we can define a non-linear iterative optimisation formulation to solve for the parameters of \(Q\) and \(E\). By differentiating the error function w.r.t. the transformation parameters gives the derivatives of the error function (\ref{eq:error_func}) and can be expressed by its Jacobian:
\[
J=
\left(
\begin{array}{cccccc}
\frac{\delta \epsilon}{\delta \theta_x} & \frac{\delta \epsilon}{\delta \theta_y} & \frac{\delta \epsilon}{\delta t_z} & \frac{\delta \epsilon}{\delta A} & \frac{\delta \epsilon}{\delta B} & \frac{\delta \epsilon}{\delta C}
\end{array}
\right),
\]

This Jacobian can be applied to the update vector \(b\) that applies an update to the transformation parameters (\(\theta_x, \theta_y, t_z, A, B, C\)) directly giving the following relationship:
\begin{equation}
J \cdot b = \epsilon \hspace{1.5em},\hspace{1.5em} b = (J^\mathrm{T} \cdot J)^{-1} \cdot J^\mathrm{T} \cdot \epsilon
\label{eq:iter_update}
\end{equation}
This provides an iterative optimisation method to solve for curvature based on the six transformation parameters in \(b\), using \(J\) and the error function (\(\epsilon\)). In order to improve the accuracy of this fitting method weights can also be added to each point of the patch. This increases the robustness to outliers, and changes equation \ref{eq:iter_update} to the following:

\begin{equation}
b = (J^\mathrm{T} \cdot W \cdot J)^{-1} \cdot J^\mathrm{T} \cdot W \cdot \epsilon
\label{eq:iter_reweighted_update}
\end{equation}

The weights are computed as a function of the error creating a robust iterative re-weighted least squares solution. This means the weights are recomputed on each iteration. Additionally the weighting function can be applied as a rejection filter to eliminate points that have an error greater than some threshold by giving them a weight of zero. The error function used in our method was applied in testing with and without the rejection filter with mixed performance. The weighting function used was:

\begin{equation}
W(\epsilon) =
\begin{cases}
k / (k + \epsilon^2), & \text{if}\ \epsilon^2 < R \\
0, & \text{otherwise}
\end{cases}
\label{eq:weighting_function}
\end{equation}

In this case \(R\) was in general twice the mean squared error of the patch and \(k\) was a constant that could be lowered or raised to be more or less aggressive to outliers respectively.

\subsubsection{Improved Normal Field}

An additional benefit of our method is that while estimating principal curvature values for each point the system can also be used to improve the initial normal estimate there. In aligning the patch to the normal of the surface and optimising for this alignment over $\theta_x$ and $\theta_y$ the system produces an improved estimate of the surface normal. Since a quadric is computed densely the entire normal field can be improved. The results of this improvement are demonstrated in section \ref{sec:normal_est}.


\section{Results}
\label{sec:results}

\subsection{Curvature Dataset}
\label{sec:dataset}

\begin{figure}[!h]
	\includegraphics[width=\linewidth]{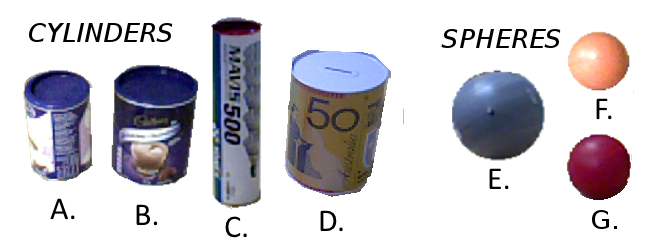}
	\caption{All objects that appear in the datasets used for real-world evaluation with measured principal curvature values. \textbf{A}:ChocA, \textbf{B}:ChocB, \textbf{C}:Badminton, \textbf{D}:Money, \textbf{E}:BallA, \textbf{F}:Stress, \textbf{G}:BallB}
	\label{fig:dataset_items}
\end{figure}

To test the accuracy of the presented metric curvature estimation method a dataset of ground-truth curvature values was created. The dataset includes both synthetic and manually labelled ground truth images captured on low-cost RGB-D sensor to test and compare the robustness of our methodology. The dataset consists of various shapes, but predominantly contains cylinders, spheres and planar surfaces. The shapes used are defined in Table \ref{table:gt_curv_prim}, along with their measured principal curvature values. Figure \ref{fig:dataset_items} shows each of the objects that appears in the dataset. Additionally points without depth measurements or those on edges and corners have been removed from consideration as they are not expected to return sensible results compared to real-world values, that is curvature is effectively infinite on edges and corners. 

{\renewcommand{\arraystretch}{1.3}%
\begin{table}[h]
	\footnotesize
	\begin{center}
		\resizebox{\linewidth}{!}{%
			\begin{tabular}{lccrccrcc}
				\multicolumn{1}{c}{} & \multicolumn{2}{c}{\textbf{Expected}} & \phantom{a} & \multicolumn{2}{c}{\textbf{Ours(It)}} & \phantom{a} & \multicolumn{2}{c}{\textbf{Error}} \\
				\cline{2-3}\cline{5-6}\cline{8-9}
				\multicolumn{1}{c}{} & \textbf{\(\kappa_2\)} & \textbf{\(\kappa_1\)} && \textbf{\(\kappa_2\)} &  \textbf{\(\kappa_1\)} && RMS & \(\sigma\) \\ 
				\multicolumn{7}{l}{\textbf{Real-World Objects}} \\  \midrule
				\multicolumn{1}{c}{\textbf{ChocA}} & 0 & 0.026 && -2.3e-4 & 0.022 && 9.1e-3 & 7.4e-3 \\  
				\multicolumn{1}{c}{\textbf{ChocB}} & 0 & 0.019 && 3.5e-4 & 0.0172 && 4.8e-3 & 3.6e-3 \\  
				\multicolumn{1}{c}{\textbf{Badminton}} & 0 & 0.029 && -9.9e-4 & 0.023 && 1.2e-2 & 8.8e-3 \\ 
				\multicolumn{1}{c}{\textbf{Money}} & 0 & 0.016 && -1.4e-4 & 0.015 && 5.5e-3 & 5.9e-3 \\ 
				\multicolumn{1}{c}{\textbf{Stress}} & 0.032 & 0.032 && 0.018 & 0.029 && 0.016 & 0.013 \\  
				\multicolumn{1}{c}{\textbf{BallA}} & 0.015 & 0.015 && 0.009 & 0.015 && 0.01 & 6.3e-3 \\ 
				\multicolumn{1}{c}{\textbf{BallB}} & 0.030 & 0.030 && 0.015 & 0.029 && 0.02 & 0.013 \\ 
				\multicolumn{7}{l}{\textbf{Synthetic Objects}} \\  \midrule
				\multicolumn{1}{c}{\textbf{Cylinder(9cm)}} & 0.0111 & 0 && 0.0111 & -1.8e-6 && 1.2e-4 & 5.8e-5 \\ 
				\multicolumn{1}{c}{\textbf{Sphere(10cm)}} & 0.010 & 0.010 && 0.010 & 0.010 && 3.7e-5 & 1.4e-4 \\
				\multicolumn{1}{c}{\textbf{Torus(10cm,3cm)}} & - & - && - & - && 6.3e-4 & 8.3e-4 \\
			\end{tabular}
		}
		\caption{Expected ground truth principal curvature values and those computed by our system averaged across all instances in each dataset. Computed from real-world objects shown in figure \ref{fig:dataset_items} and synthetic objects shown in figure \ref{fig:per_item_results_synth}.}
		\label{table:gt_curv_prim}
	\end{center}
\end{table}

\subsection{Measuring System Accuracy}
\label{sec:system_accuracy}

Several tests were performed to accurately measure the accuracy of the presented system on real-world and synthetic data. We contrast our method to several implementations, including the least-squares method (\textbf{Douros}) used in \cite{Douros2002}, Re-weighted least-squares(\textbf{Besl}) from \cite{Besl1988} and the method used in Point Cloud Library (\textbf{PCL}) compared to our proposed method (\textbf{Ours}) and a variant that includes rejection (\textbf{Ours+R}).

\begin{figure}[h!]
	\begin{center}
		\includegraphics[width=\linewidth]{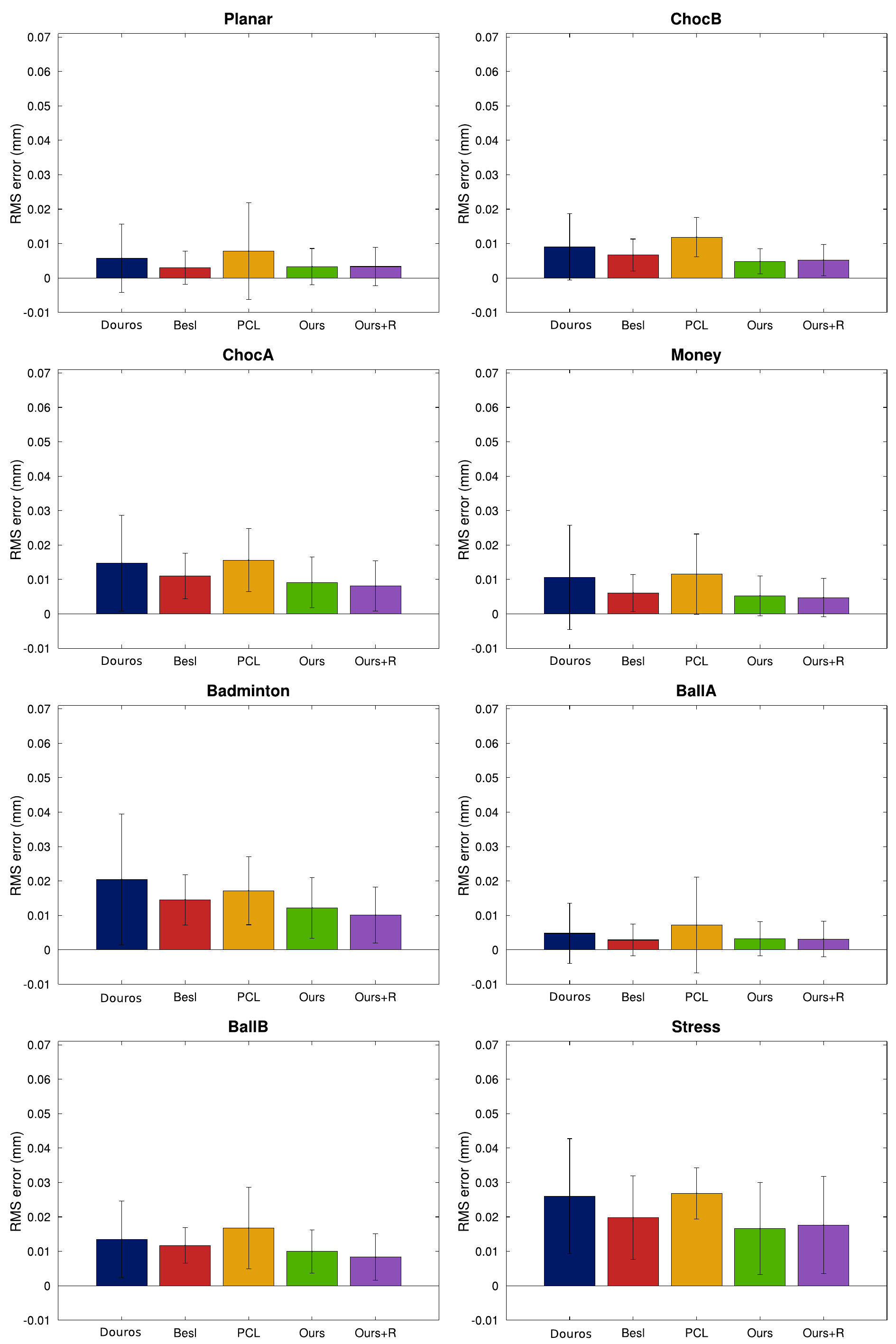}
		\caption{RMS error in estimating principal curvature on real-world data of each object in the dataset for several methods. Least squares (\textbf{Douros}), Re-weighted Least Squares (\textbf{Besl}), Point Cloud Library w. 30mm radius (\textbf{PCL}), Our Iterative Method (\textbf{Ours}) and Our Iterative Method with a rejection filter (\textbf{Ours+R}). This comparative test shows the improvement in our method.}
		\label{fig:per_item_results_real}
	\end{center}
\end{figure}

\begin{figure}[h]
	\begin{center}
		\includegraphics[width=\linewidth]{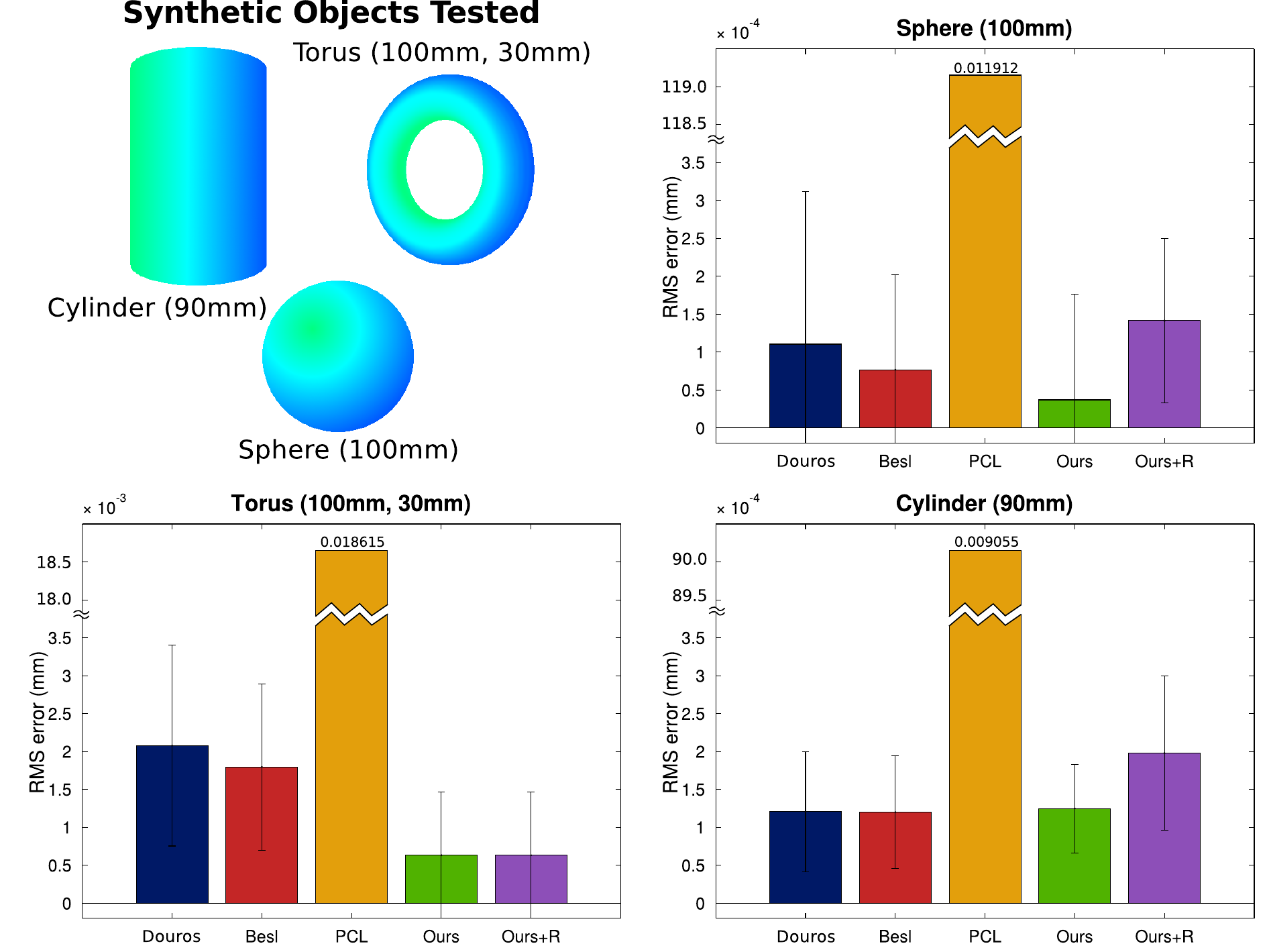}
		\caption{RMS error in estimating principal curvature on synthetic objects for several methods. Least squares (\textbf{Douros}), Re-weighted Least Squares (\textbf{Besl}), Point Cloud Library with a 10mm radius (\textbf{PCL}), Our Iterative Method (\textbf{Ours}) and Our Iterative Method with a rejection filter (\textbf{Ours+R}).}
		\label{fig:per_item_results_synth}
	\end{center}
\end{figure}

\subsubsection{RMS error for Real-world and Synthetic Scenes}
\label{sec:rms_error}

The RMS error of the estimated principal curvature values collected from synthetic and real-world datasets was compared for our method and several other methods, including curvature estimation from quadrics such as that used in \cite{Douros2002} (\textbf{Douros}) and \cite{Besl1988} (\textbf{Besl}), the method used by PCL and the two variants produced for this paper. The computed curvature values for each object as shown in Table \ref{table:gt_curv_prim} and demonstrate the relative accuracy of our system. The results of testing across all scenes are summarised in figures \ref{fig:per_item_results_real} and \ref{fig:per_item_results_synth} and demonstrate the improvement in accuracy of our method (\textbf{Ours}) over those tested. Our method produces consistently lower RMS error across each of the objects considered in all scenes. The curvature was estimated for PCL using a KD-tree search neighbourhood radius of 20mm for real-world and 10mm for synthetic. The choice to use the 30mm radius was made based on a several tests which shows this radius was consistently the best choice for the data. This also corresponds to a patch roughly the same size as the patch used in the other methods. All remaining methods use the same sparsely sampled patch of \(37\times37\) values in view space.

\begin{figure}[h!]
	\begin{center}
		\includegraphics[width=0.80\linewidth]{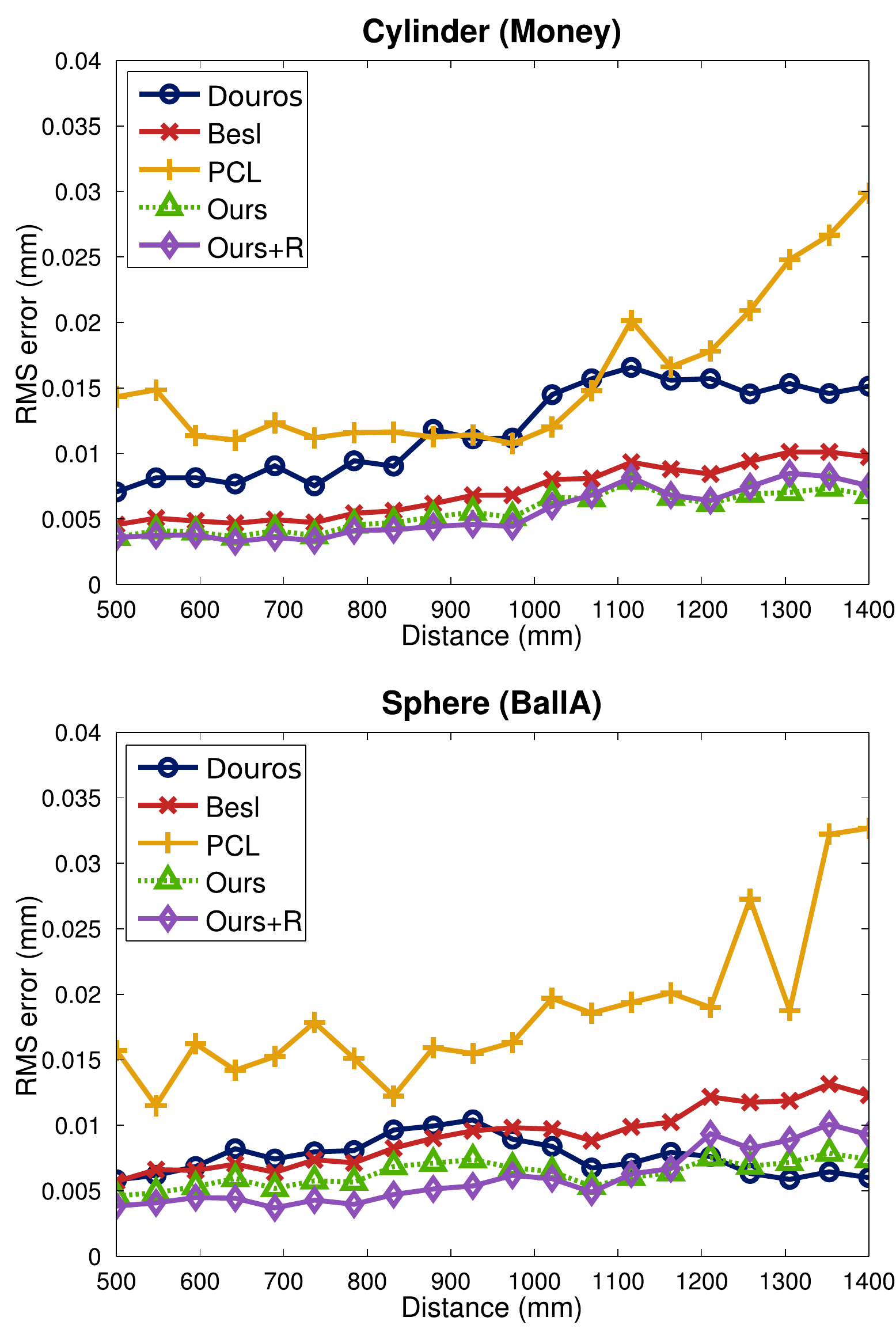}
		\caption{RMS error in estimating principal curvature on two real-world scenes where the object of interest was incrementally positioned further and further from the device to investigate how consistent the principal curvature measurements are for the several methods compared.}
		\label{fig:curv_err_dist}
	\end{center}
\end{figure}

\subsubsection{Error vs Object Distance}
\label{sec:err_distance}

As our method uses a fixed size patch for estimation which may create concerns about consistency over distance from the camera.  In order to measure the consistency of the system over large changes in distance we created two datasets from real-world objects. Figure \ref{fig:curv_err_dist} shows the results of testing the consistency of the curvature measurements from our method and several others while repeatedly moving an object further away from the camera. This motion is expected to increase the level of quantisation noise in the camera. An increase in error is experienced in all cases, however our method produces the lowest errors in almost all cases. PCL performs worse in this regard, partly due to the fixed radius sphere used in the k-d tree computation and could be improved with a more complex strategy for choosing the search radius.   

\subsubsection{Error vs Noise}
\label{sec:err_noise}

Robustness to noise was tested for each method by successively increasing surface noise on a synthetically generated sphere while estimating its principal curvature values. The result of this test is shown in Figure \ref{fig:curv_err_noise}. Due to the aggressive way the noise was added all methods are effected by the added noise. While the quadric based methods provide considerably better estimates for the curvature at each noise level, with PCL consistently underestimating the principal curvatures. Again our method out performs the others in terms of robustness to noise.

\begin{figure}[h!]
	\begin{center}
		\includegraphics[width=0.8\linewidth]{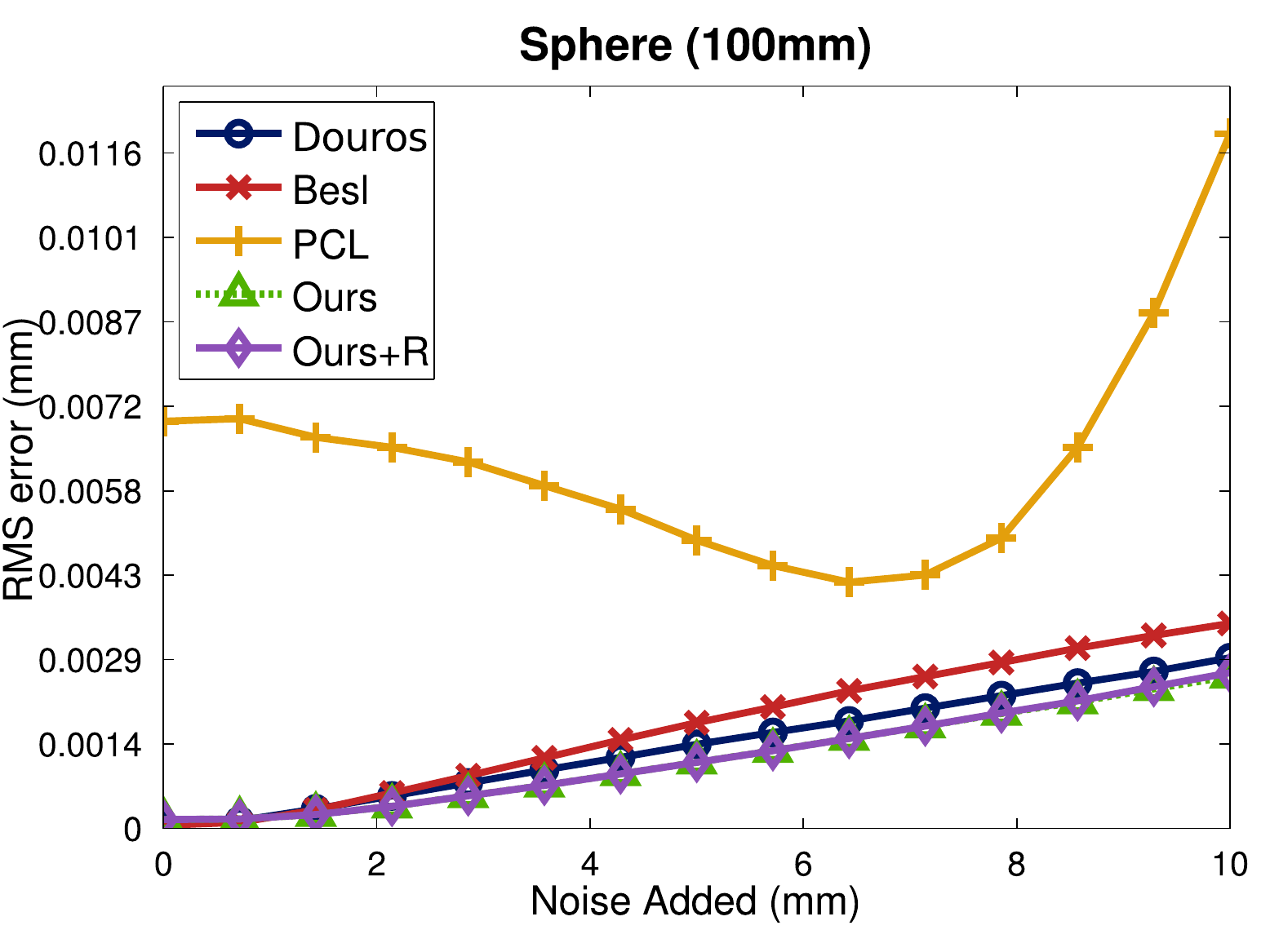}
		\caption{RMS error comparision for several methods in estimating principal curvature on a synthetically generated sphere (100mm radius) with increasing random noise added. Iteratively (\textbf{Ours}) and Iteratively w. Rejection (\textbf{Ours+R}) are the methods developed for this paper.}
		\label{fig:curv_err_noise}
	\end{center}
\end{figure}

\subsection{Refined Normal Estimatation}
\label{sec:normal_est}

A qualitative comparison of normal estimations is shown in Figure \ref{fig:normal_est}. Part of our computation requires the alignment of the quadric surface to fit the points and this is achieved by rotating the surface and altering its curvature. The rotation applied to the quadric surface during the fitting stage can give a more accurate indication of the true direction of the surface normal than the original estimate. Figure \ref{fig:normal_est} demonstrates this improvement where \textbf{A} shows a comparison to the normals computed by the PCL library and \textbf{B} shows the refined normals produced by our method. 

A quantitative comparison of the normal estimation robustness was also performed on a synthetically generated sphere, where the known normals were compared for mean absolute error in theta (angle between the normals). The result is shown in figure \ref{fig:norm_err}. This demonstrates that our system is able to produce smoother and more accurate estimates of normals, than surface differentiation via principal component analysis alone.

\begin{figure}[h!]
	\begin{center}
		\includegraphics[width=\linewidth]{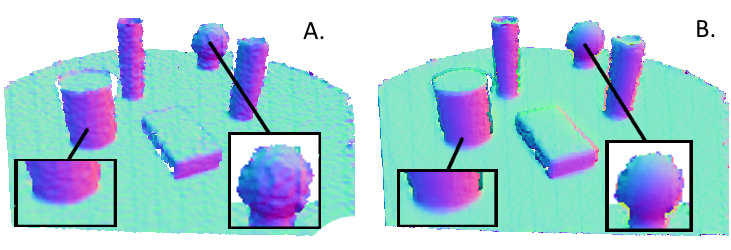}
		\caption{Qualitative comparison of normal estimation from real-world data. \textbf{A:} Normals estimated using PCA of points differences in a sphere of radius 10mm. \textbf{B:} Refined normal estimates from the quadric surface patch calculated in our method. Note that the boundaries still remain sharp even with the added smoothness of the normals.}
		\label{fig:normal_est}
	\end{center}
\end{figure}

\begin{figure}[h!]
	\begin{center}
		\includegraphics[width=0.8\linewidth]{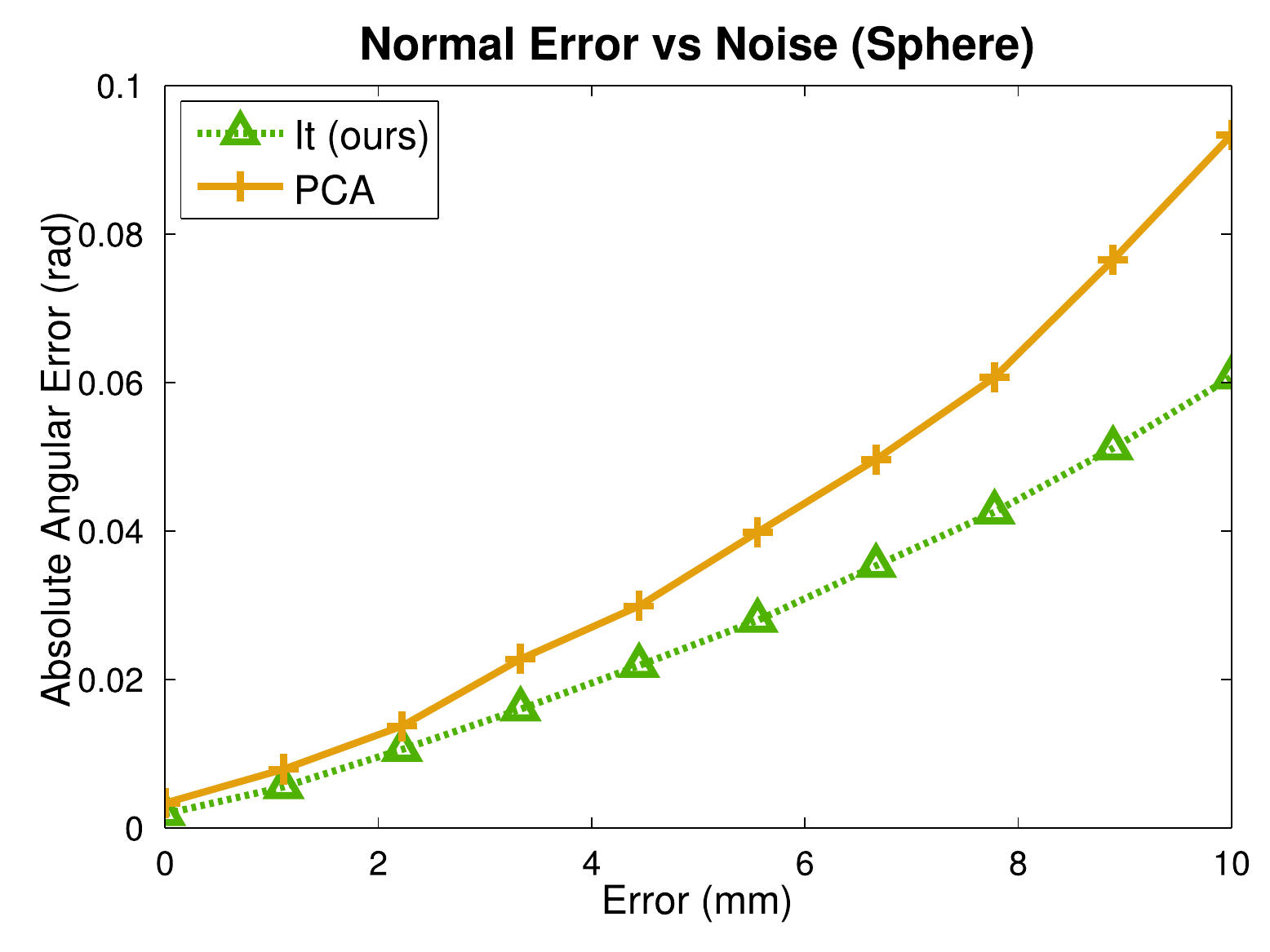}
		\caption{Comparison of mean absolute angular error in estimated normals of a synthetic sphere with increasing surface noise (in mm) for our method (\textbf{It}) and using PCA over the surface such as that used in PCL and many others. This shows an iterative method such as our can perform more robustly in the presence of noise when estimating normals.}
		\label{fig:norm_err}
	\end{center}
\end{figure}

\begin{table}[h!]
	\begin{center}
		\resizebox{.9\linewidth}{!}{%
			\begin{tabular}{ crccc }
				\multicolumn{1}{c}{} & \phantom{a} & \multicolumn{3}{c}{\textbf{Patch Size(pixels)}} \\
				\cline{3-5}
				\textbf{Image Size(pixels)} && \(9\times 9\) & \textbf{\(21\times 21\)} & \textbf{\(37\times 37\)} \\  
				\cline{1-1}\cline{3-5}
				\textbf{640x480} && 10.2ms & 16.1ms & 19.3ms \\ 
				\textbf{512x424} && 6.7ms & 12.2ms & 13.2ms \\ 
				\textbf{320x240} && 2.5ms & 4.5ms & 5.0ms \\ 
			\end{tabular}
		}
		\caption{Timing performance of curvature estimation for our iterative method averaged over several runs at each scale for each patch size.}
		\label{table:timing}
	\end{center}
\end{table}

\subsection{Timing}
\label{sec:timing}

Our curvature estimation method was tested on a system with a single NVIDIA K40c GPU. The results of timing our system are summarised in table \ref{table:timing}. This shows that the system operates comfortably at frame rate (50-60Hz) even for large patch sizes \(37\times37\). The \(37\times37\) patch was used in all experimental results as it allows the highest accuracy and resolution for the given data. However with more accurate data the smaller patch sizes could be used allowing for faster computation times.

\definecolor{it_green}{rgb}{0.31, 0.70, 0}

\newcommand{\DoStuff}[1]{%
	\pgfmathsetmacro{\PercentColor}{100*(#1)}
	\xdef\PercentColor{\PercentColor}%
	\cellcolor{it_green!\PercentColor!white}\pgfmathprintnumber[int trunc]{\PercentColor}
}

\newcommand{\IfNumber}[1] {
	\IfDecimal{#1}
	{\DoStuff{#1}}
	{#1}%
}

\newcolumntype{L}{>{\collectcell\IfNumber}l<{\endcollectcell}}


\begin{table}[h]
	\begin{center}
		\resizebox{.95\linewidth}{!}{%
			\begin{tabular}{lLLLLLLLLL}
				\multicolumn{1}{r}{} & \multicolumn{1}{r}{\rotatebox{90}{Planar}} & \multicolumn{1}{r}{\rotatebox{90}{Background}} & \multicolumn{1}{r}{\rotatebox{90}{ChocA}} & \multicolumn{1}{r}{\rotatebox{90}{ChocB}} & \multicolumn{1}{r}{\rotatebox{90}{Badminton\hspace{0.3em}}} & \multicolumn{1}{r}{\rotatebox{90}{Money}} & \multicolumn{1}{r}{\rotatebox{90}{Stress}} & \multicolumn{1}{r}{\rotatebox{90}{BallA}} & \multicolumn{1}{r}{\rotatebox{90}{BallB}}\\ 
				ChocA & 0.12 & 0.16 & 0.49 & 0.04 & 0.25 & 0.01 & 0.01 & 0.00 & 0.01 \\
				ChocB & 0.05 & 0.14 & 0.11 & 0.67 & 0.00 & 0.00 & 0.01 & 0.01 & 0.00 \\ 
				Badminton & 0.14 & 0.18 & 0.25 & 0.00 & 0.37 & 0.03 & 0.00 & 0.00 & 0.02 \\ 
				Money & 0.14 & 0.15 & 0.05 & 0.00 & 0.06 & 0.58 & 0.00 & 0.01 & 0.00 \\ 
				Stress & 0.05 & 0.53 & 0.13 & 0.06 & 0.00 & 0.00 & 0.23 & 0.00 & 0.00 \\ 
				BallA & 0.05 & 0.10 & 0.05 & 0.00 & 0.02 & 0.07 & 0.00 & 0.68 & 0.01 \\ 
				BallB & 0.04 & 0.19 & 0.04 & 0.00 & 0.05 & 0.04 & 0.00 & 0.00 & 0.64 \\
			\end{tabular}
		}
	\end{center}
	\caption{The confusion matrix for our iterative method demonstrating our method can be used to locate corresponding regions of known or similar curvature across multiple views.}
	\label{table:confuseGPU}
\end{table}

\begin{table}[h!]
	\begin{center}
		\resizebox{.95\linewidth}{!}{%
			\begin{tabular}{lLLLLLLLLL}
				\multicolumn{1}{r}{} & \multicolumn{1}{r}{\rotatebox{90}{Planar}} & \multicolumn{1}{r}{\rotatebox{90}{Background}} & \multicolumn{1}{r}{\rotatebox{90}{ChocA}} & \multicolumn{1}{r}{\rotatebox{90}{ChocB}} & \multicolumn{1}{r}{\rotatebox{90}{Badminton\hspace{0.3em}}} & \multicolumn{1}{r}{\rotatebox{90}{Money}} & \multicolumn{1}{r}{\rotatebox{90}{Stress}} & \multicolumn{1}{r}{\rotatebox{90}{BallA}} & \multicolumn{1}{r}{\rotatebox{90}{BallB}}\\ 
				ChocA & 0.36 & 0.38 & 0.12 & 0.03 & 0.06 & 0.05 & 0.00 & 0.05 & 0.03 \\ 
				ChocB & 0.55 & 0.26 & 0.07 & 0.09 & 0.0 & 0.00 & 0.00 & 0.03 & 0.00 \\
				Badminton & 0.25 & 0.39 & 0.08 & 0.00 & 0.11 & 0.05 & 0.00 & 0.07 & 0.04 \\
				Money & 0.56 & 0.23 & 0.05 & 0.00 & 0.03 & 0.08 & 0.00 & 0.02 & 0.01 \\ 
				Stress & 0.22 & 0.51 & 0.05 & 0.02 & 0.00 & 0.00 & 0.10 & 0.06 & 0.00 \\ 
				BallA & 0.42 & 0.33 & 0.05 & 0.01 & 0.05 & 0.03 & 0.00 & 0.06 & 0.03 \\ 
				BallB & 0.08 & 0.28 & 0.06 & 0.00 & 0.09 & 0.03 & 0.00 & 0.08 & 0.36\\
			\end{tabular}
		}
		\caption{The confusion matrix for PCL demonstrating a sensitivity to noise and lack of discriminative power, with many of the points matching outside the desired regions.}
		\label{table:confusePCL}
	\end{center}
\end{table}

\subsection{Curvature-based Correspondence Estimates}
\label{sec:curv_correspond}

The consistency of our systems principal curvature estimation was compared to PCL using point correspondence estimates on real-world data over different viewpoints. We used the CPU PCL implementation, using a K-D tree for the search space with search neighbourhood radius of 20mm and compared the matching performance to our system with a patch size of \(37\times37\) (sampled sparsely).

The correspondence test was conducted as follows:
\begin{itemize}
	\item A sample point is selected inside each object for every frame in a dataset.
	\item The closest 400 curvature values to every sample point are computed for each image in the series.
	\item Each of the closest values will belong to an object in the dataset, part of the table plane or fall on the unclassified region (Background). 
	\item The number of values for each object forms a histogram for each object which shows how consistently each test point matches across multiple views. The ideal case is all points belong to the sample point object.
\end{itemize}

The confusion matrices in tables \ref{table:confuseGPU} and \ref{table:confusePCL} demonstrate the result of this testing for our method and PCL. These confusion matrices indicate the percentage for each object each of the selected objects matched to. The results in table \ref{table:confuseGPU} clearly indicate that our system is able to consistently estimate the curvature across different viewpoints and images. This makes it a viable method for estimating corresponding points across a wide-baseline. The results of this test also indicates that smaller objects with high curvature are the most difficult (such as the stress ball) to distinguish for our system and objects with similar curvature profiles will also be more easily confused (such as chocA and badminton). The results for PCL in table \ref{table:confusePCL} indicate the relatively poor consistency of their curvature estimation. PCL consistently estimates correspondences to be outside of classified objects (Background). The results of this experiment demonstrates our system could be used to aid object detection or recognition by drastically reducing the size of the search space for correspondence estimation.


\section{Conclusions and Future Work}

The system demonstrated in this work is an accurate real-time GPU based curvature estimator for noisy point cloud data, generated from low-cost RGB-D sensors, which operates at frame rate for Kinect style RGB-D data. The applications of this work are varied, including segmentation, normal field calculation and correspondence estimation. The results of this work demonstrates the viability of this approach to robotics applications including object shape matching and inferring object characteristics including size estimates and as a wide-baseline correspondence estimation tool.

An intended extension of this work is to apply it to object segmentation of scenes \cite{Besl1988} \cite{Alshawabkeh2008} and object recognition and classification. Preliminary testing has shown this to be applicable to solving these tasks.
	

\section{Acknowledgements}

This work was supported by the Australian Research Council Centre of Excellence for Robotic Vision (project number CE140100016).


%
%


\end{document}